\documentclass{llncs}
\usepackage[american]{babel}
\usepackage[T1]{fontenc}
\usepackage{times}
\usepackage{graphicx}
\usepackage{authblk}
\usepackage{color}

\begin{document}

\title{The Case for Being Average: A Mediocrity Approach\\ to Style Masking and Author Obfuscation}
\subtitle{(Best of the Labs Track at CLEF-2017)}

\author{Georgi Karadzhov\inst{1} \and Tsvetomila Mihaylova\inst{1} \and Yasen~Kiprov\inst{1} \and Georgi Georgiev\inst{1} \and Ivan~Koychev\inst{1} \and Preslav Nakov\inst{2}}
\institute{Faculty of Mathematics and Informatics, Sofia University ``St. Kliment Ohridski'', Bulgaria \\
\email{\{georgi.m.karadjov, tsvetomila.mihaylova\}@gmail.com, \{yasen.kiprov, g.d.georgiev\}@gmail.com, koychev@fmi.uni-sofia.bg}
\and
Qatar Computing Research Institute, HBKU, Qatar \\
\email{pnakov@qf.org.qa}
}

\maketitle

\begin{abstract}
Users posting online expect to remain anonymous unless they have logged in, which is often needed for them to be able to discuss freely on various topics. Preserving the anonymity of a text's writer can be also important in some other contexts, e.g., in the case of witness protection or anonymity programs. However, each person has his/her own style of writing, which can be analyzed using stylometry, and as a result, the true identity of the author of a piece of text can be revealed even if s/he has tried to hide it. Thus, it could be helpful to design automatic tools that can help a person obfuscate his/her identity when writing text. In particular, here we propose an approach that changes the text, so that it is pushed towards average values for some general stylometric characteristics, thus making the use of these characteristics less discriminative. The approach consists of three main steps: first, we calculate the values for some popular stylometric metrics that can indicate authorship; then we apply various transformations to the text, so that these metrics are adjusted towards the average level, while preserving the semantics and the soundness of the text; and finally, we add random noise. This approach turned out to be very efficient, and yielded the best performance on the Author Obfuscation task at the PAN-2016 competition.
\end{abstract}

\section{Introduction}

An important characteristic of the Web nowadays is the perceived anonymity of online activity. For example, a user posting in a forum online would expect to remain anonymous unless the site has asked him/her to share personal information as part of the registration to use their services. This is in theory. The reality is that there is very little anonymity online as Web sites track users in various ways, e.g., by requiring registration, by using cookies, by using third-party services, by linking phone numbers or Google/Facebook accounts to online activity, etc. 
One could try to gain some anonymity by creating a new account or by using a different device or even an anonymous proxy service.
Yet, when posting in a forum, the author's anonymity is still potentially at risk, as it is possible to analyze and match his/her posts to those of a known user.
This is because each person has his/her own style of writing, which reflects their personality. 

\noindent Revealing a person's identity requires (\emph{i})~a hypothesis about who that person might be and (\emph{ii})~a sufficient sample of text written by that person. Then, stylometric features can be used to predict whether the author of a target piece of text is indeed the one hypothesized. Even without a hypothesis about a target author, stylometry can reveal key demographic characteristics about the author of a piece of text, e.g., his/her gender and age, which is of interest to marketing analysts, political strategists, etc.

Overall, stylometry is a well-established discipline with application to authorship attribution, plagiarism detection, author profiling, etc. However, much less research has been done on the topic of author obfuscation, i.e., helping a person hide his/her own style in order to protect his/her identity and key demographic information. Unlike authorship attribution or author profiling, this is not a simple text classification problem but rather a complex text generation task, where not only the author's style has to be hidden, but the text needs to remain grammatically correct and the original meaning has to be preserved as much as possible.

Below we focus on the task of author obfuscation, i.e., given a document, the goal is to paraphrase it, so that its writing style does not match the style of its original author anymore. We use the task formulation, the data, and the evaluation setup from PAN'2016, the 15th evaluation lab on uncovering plagiarism, authorship, and social software misuse.

A system addressing the task needs to optimize three conflicting objectives simultaneously:


\begin{enumerate}
\item \textbf{Safety:} forensic analysis should not be able to reveal the original author of the obfuscated text;
\item \textbf{Soundness:} the obfuscated text should be semantically equivalent to the original;
\item \textbf{Sensibility:} the obfuscated text should be inconspicuous, i.e., it should not raise suspicion that there has been an attempt to obfuscate the author.
\end{enumerate}

The performance of a participating system in the PAN'2016 task is measured as follows:
\begin{itemize}
\item \emph{automatically}: using automatic authorship verifiers to measure \emph{safety};
and
\item \emph{manually}: using peer-review to assess \emph{soundness} and \emph{sensibility}.
\end{itemize}


As these objectives are conflicting, the task is very challenging. While hiding author's style is a nontrivial task by itself, producing inconspicuous output is where most systems actually fail. At PAN'2016 \cite{mihaylova:2016}, we proposed a method for author obfuscation that performed best in terms of \emph{safety}, but lagged behind in \emph{sensibility}. One of the other systems \cite{mansoorizadeh:2016} generated text that scored high in terms of \emph{sensibility} and \emph{soundness}, but their system performed poorly in terms of \emph{safety}. Below we describe and we evaluate both our original approach and a modification thereof that addresses the issue with poor results in \emph{sensibility}. The evaluation results on the PAN'2016 dataset demonstrate the potential of our approach.

The remainder of this paper is organized as follows: Section~\ref{sec:related} introduces related work. Section~\ref{sec:method} describes our method: both the original one from PAN'2016 and the above-described modification thereof. Section~\ref{sec:evaluation} presents the evaluation setup and discusses the results. Finally, Section~\ref{sec:future} concludes and points to some possible directions for future work.

\section{Related Work}
\label{sec:related}


\paragraph{Author Identification}
is well-studied and has a long history  \cite{oldresearch10.2307/1764604}.
The most common approach is to use  variety of features \cite{holmes1998evolution,Juola2011,potthast:2016,Stamatatos:2009:SMA:1527090.1527102} such as \emph{punctuation} (e.g., relative frequency of commas), \emph{lexical} (e.g.,~frequency of function words, average word and sentence length, etc.), \emph{character-level} (e.g.,~$n$-gram frequencies), \emph{syntactic} (e.g.,~frequencies of different parts of speech), \emph{semantic} (e.g.,~frequency of various semantic relations), and \emph{application-dependent} (e.g.,~style of writing email messages).

Author Identification has been a task at the PAN competition\footnote{\url{http://pan.webis.de}} 
since 2011. The most commonly-used features at PAN'2015 \cite{Stamatatos2015OverviewOT} 
include the lengths of words, sentences, and paragraphs, type-token ratios, the use of \emph{hapax legomena} (i.e.,~words occurring only once),
character $n$-grams (including unigrams), words, punctuation marks, stopwords, and part of speech (POS) $n$-grams. Other features that some participants used analyze the text more deeply by checking style and grammar. 

Another line of research for author identification uses neural networks to induce features automatically. For example, Bagnal \cite{bagnall2015author} used a recurrent neural network based on character unigrams, thus building a character-level language model. The idea is that a language model trained on texts by a particular author will assign higher probability to texts by the same author compared to texts written by other authors.

\paragraph{Author Obfuscation.}
Research in author obfuscation has explored manual, computer-aided, and automated obfuscation \cite{potthast:2016}.
For manual obfuscation, people have tried to mask their own writing style as somebody else's, which was shown to work well \cite{almishari2014fighting,Brennan:2012,brennan2009practical}.
Computer-aided obfuscation uses tools that identify and suggest parts of text and text features that should be obfuscated, but then the obfuscation is to be done manually \cite{Kacmarcik:2006,Le:2015:SOA:2977360.2977368,McDonald:2012}.

Kacmarcik et al. \cite{Kacmarcik:2006} explored author masking by detecting the most commonly-used words by the target author and then trying to change them. They also mention the application of machine translation as a possible approach to author obfuscation. 
Other authors also used machine translation for author obfuscation \cite{Brennan:2012,export:69170}, e.g., by translating passages of text from English to one or more other languages and then back to English.
Brennan et al. \cite{Brennan:2012} investigated three different approaches to adversarial stylometry: obfuscation (masking author style), imitation (trying to copy another author's style), and machine translation. They further summarized the most common features people used to obfuscate their own writing style. 
  
Juola et al. \cite{Juola2011} developed a complex system for author obfuscation which consists of three main modules: canonization (unifying case, normalizing white spaces, spelling correction, etc.), event set determination (extraction of events significant for author detection, such as words, parts of speech bi- or tri-grams, etc.), and statistical inference (measures that determine the results and confidence in the final report). The authors used this same approach \cite{Juola:2012:DSD:2388616.2388630} to detect deliberate style obfuscation. 


Kabbara et al. \cite{kabbara2016stylistic} used long short-term memory recurrent neural networks to transform the text in a similar fashion as in machine translation, but essentially `translating' from one author's style to the style of other authors. While the approach looks promising, there was no proper evaluation in terms of safety and soundness.

\section{Our Mediocrity Approach to Style Masking and Author Obfuscation}
\label{sec:method}



The main idea behind our approach is to measure the most significant features of the text used for author identification as mentioned in the work of Brennan et al. \cite{Brennan:2012}. Then, we apply transformations of the text, so that the values of these metrics are pushed towards average.

Our approach consists of three main steps. First, we calculate ``average'' metrics based on the training corpus provided for the PAN-2016 Author Obfuscation task \cite{potthast:2016} and a corpus of several public-domain books from Project Gutenberg.\footnote{We used the following books: \emph{The Adventures of Sherlock Holmes} by Sir Arthur Conan Doyle,
\emph{History of the United States} by Charles A. Beard and Mary R. Beard, \emph{Manual of Surgery Volume First: General Surgery} by Alexis Thomson and Alexander Miles. Sixth Edition., and \emph{War and Peace}, by Leo Tolstoy.} 
We will call these average values the \textit{calculated averages}, and we will try to push the document metrics towards them.
Second, we calculate the corresponding metrics for each target document. Then we apply ad hoc transformations on the text aiming to average those metric for the document.
Third, we apply additional transformations aiming to randomly change the average metrics of the text. At this step, we apply very harmless transformations, so that we can preserve the meaning as much as possible. For example, we use dictionaries to transform abbreviations, equations, and short forms to listed alternatives.

\subsection{Calculating Text Metrics}

We calculate the following text metrics:
\begin{enumerate}
\item Average number of word tokens per sentence;
\item Punctuation to word token count ratio;
\item Stop words to word token count ratio;
\item Word type to token ratio;
\item POS to word count ratio: measured for four part-of-speech groups: nouns, verbs, adjectives, and adverbs; 
\item Uppercase word tokens to all word tokens count ratio;
\item Number of mentions of each word type in the text.
\end{enumerate}
Given a document to be obfuscated, we calculate the above measures for it.
Then, we split the document into parts, and for each part, we compare the measures for this part to the document-level averages. Finally, we apply transformations to push these values towards the corpus-level average.

\subsection{Modulizing the Text}

We used the PAN-2016 Author Obfuscation task setup, i.e., each text was to be split into parts of up to 50 words each. To do this, we first segmented the text into sentences using the NLTK sentence splitter. Then, we merged some of these sentences to get bigger segments, while keeping the segment lengths under 50 words. We ignored paragraph boundaries in this process.

\subsection{Text Transformations}
\begin{enumerate}
\item \textbf{Splitting or merging sentences}

If the average sentence length of the whole document is below the \textit{calculated average} of this metric,
we perform merging of the sentences for each text part. We merge all the sentences for a given text part into one sentence. Merging is done by adding a random connecting word (\textit{and, as, yet}) and randomly inserting punctuation - comma (,) or semicolon (;).
When the average sentence length of the entire document is above the average, we split the sentence into shorter ones. We use a simple sentence splitting algorithm: we go through all POS-tagged words in the text, we count the nouns and the verbs, and when we reach a conjunction \textit{and}, if the sentence so far contains a noun and a verb, we replace the \textit{and} with a full stop (.) and we capitalize the next word as it will now start a new sentence.

\item \textbf{Stop Words} 

Stop words can be strong indicators for author identification as people have the tendency to use and overuse specific stop words. Thus, we perform two kinds of transformations regarding stop words:
\subitem removing stop words that carry little to no information;
\subitem replacing stop words with alternatives or with a phrase with the same meaning.

\item \textbf{Spelling} 

The spelling score of a document is high if there are no spelling mistakes, and low when there are some. 
\subitem If we need to increase the spelling score, we apply automatic spelling correction. Our spell-checker uses a probabilistic model and a previously mentioned set of publicly-available books.
\subitem If we need to decrease the score, we use a dictionary to insert common mistakes in the text. The 
dictionary was manually created using data from various sources. 

\item \textbf{Punctuation} 

If the punctuation use is above average, we remove some punctuation within the sentence. This is limited to the symbols comma (,), semicolon (;), and colon (:)
If the punctuation use is below average, we apply the following two techniques to improve that score:
\subitem We randomly insert comma or semicolon before prepositions. We insert comma with a higher probability compared to semicolon.
\subitem We insert redundant symbols using the following schema:
\begin{verbatim}
! can be replaced with !, !!, or !!!
? can be replaced with ?, ??, ???, ?!?, or !?!
\end{verbatim}

\item \textbf{Word Substitution}

In order to change the frequency of word types, we replace the most or the least common words. We use synonyms, hypernyms or word descriptions from WordNet \cite{wordnet,Miller:1995:WLD:219717.219748}. In particular, if the document type-token ratio is above average, we replace the most frequently used words in the document 
by random synonyms or hypernyms.
In contrast, if the ratio is below average, we randomly replace the least frequently used words 
with their definitions.

\item \textbf{Paraphrase Corpus}

We randomly replace phrases from the text with variants from a paraphrase corpus. In particular, we use the short version of the phrasal corpus of PPDB, the Paraphrase Database (\cite{ganitkevitch2013ppdb}). 
As a result, the meaning of the text is still preserved, but there is improvement for the metrics for individual word frequencies and parts of speech.

\item \textbf{Uppercase Words}

If we need to decrease the proportion of uppercase words, we lowercase words that are all in uppercase and contain at least four letters. We assume that if a word is in uppercase and is less than four letters long, it is likely to be an acronym, and thus we should keep it in uppercase. 
\end{enumerate}


\subsection{Noise}

Having applied the above transformations, we then insert some random noise in the text, using the following two operations:

\begin{enumerate}

\item \textbf{Switching British and American English }

We randomly change words from British to American English and vice versa, using a lexicon.\footnote{We have released our code, including all our lexicons, in the following repository: \url{https://bitbucket.org/pan2016authorobfuscation/authorobfuscation/}}

\item \textbf{Inserting random functional words}

We insert random functional words at the beginning of the sentence. The words are taken from a discourse marker lexicon.

\end{enumerate}

\subsection{General Transformations}
We also apply some general transformations that preserve the meaning of the text while helping mask the author style.

\begin{enumerate}
\item \textbf{Replacing short forms}

We replace short forms such as \textit{I've, I'd, I'm, I'll, don't, etc.} with their full forms.

\item \textbf{Replacing numbers with words}

We replace tokens that are POS-tagged as numbers with their word representation in English.

\item \textbf{Replacing equations}

As there were some examples of scientific text in the training corpus, if the text contains equations, the operations in them are being replaced with words. An equation is recognized if the text contains both comparison and inner equation symbols: 
\begin{verbatim}
".[<>=]+." and ".[\+\-\*\/]+."
\end{verbatim}

We replace the following symbols if we find an equation: \textit{$+$~(plus), $-$~(minus), $*$~(multiplied by), $/$~(divided by), $=$~(equals), $>$~(greater than), $<$~(less than), $<=$~(less than or equal to), $>=$~(greater than or equal to)}.

\item \textbf{Replace symbols and abbreviations with words}

We further replace the following symbols and abbreviations with their word representations: currency symbols, \textit{\% (percent), @ (at)}, abbreviations of person titles (such as \textit{Prof.,  Mr., Dr.}, etc.).

\item \textbf{Simple transformations with regular expressions}

Possessions (genitive markers) are replaced by a shorter form (e.g., \textit{book of John} becomes \textit{John's book}); here is the corresponding regular expression:
\begin{verbatim} "(\w+) of (\w+)"  is replaced with  "\2's \1" \end{verbatim}
This will slightly change the stop words rate and could also obfuscate writing in general as the former way of expressing possession is generally less common.
\end{enumerate}

\subsection{Other Methods}

\begin{enumerate}
\item \textbf{Machine translation}

We also experimented with machine translation as described in \cite{Brennan:2012}. 
We used the \textit{Microsoft Translator API} to translate from English to Croatian and Estonian, and then back to English.
The general assumption is that by applying machine translation using different languages, we will naturally paraphrase the text, while preserving the meaning.
However, our manual evaluation has shown that this often yielded text whose meaning differed from that of the original text.
This is consistent with the observations of Keswani et al. \cite{keswani:2016}, who achieved poor sensibility and soundness by cyclic translations through several languages and then back to English for their PAN-2016 system \cite{potthast:2016}.
 
\item \textbf{Simple word substitution}

Another way to approach the task is to perform simple word substitution, e.g., using WordNet. 
Unfortunately, substituting a word does not always yield fluent text due to grammatical (e.g.,~wrong word inflection or wrong part of speech) or semantic mismatch (i.e.,~even though the substituted word may be a good paraphrase in general, it is not a good fit in the particular context). 
The problem can be alleviated to some extent by using multi-word paraphrases as longer phrases are less ambiguous and thus less context-dependent.
In any case, the result of word/phrase substitution is not perfect; yet, it was found to perform much better than using machine translation in terms of sensibility and soundness. In particular, the two approaches were compared in the PAN-2016 Author Obfuscation task \cite{potthast:2016}, where a substitution system outperformed a system based on round-trip machine translation \cite{keswani:2016} in terms of both soundness and sensibility; however, both systems scored low on safety.
Mansoorizadeh et al. \cite{mansoorizadeh:2016} performed substitution in a different manner, achieving high sensibility and soundness scores; however, their system was the worst in terms of safety in the PAN-2016 Author Obfuscation task. 
We tried their approach, and after some initial experiments, we concluded that word/phrase substitutions should be used not in isolation but rather together with our above-described transformations, which limits the use of such substitutions to some specific cases \cite{mihaylova:2016}. 

\end{enumerate}

\subsection{Transformation magnitude}

After reviewing the results for the system we submitted to PAN-2016 \cite{mihaylova:2016}, we noticed that in some cases our transformations were too aggressive. 
In particular, if the value of some metric was below the average, we applied transformations to increase it, but we did not have a mechanism to control by how much we were boosting it. This sometimes resulted in undesired behaviour, e.g.,~when the value of a metric was close to the average, we could over-push it significantly over/below the average. Effectively, this goes against our aim to push it towards the average. 
As a side effect, in some cases, the text readability was affected negatively as well.
In order to address the issue, we introduced an additional parameter, which tracks the magnitude of the desired change, i.e.,~the difference between the current value and the average value. 
We further modified the above transformations to keep track of and to update the value of this parameter accordingly.

\section{Evaluation}
\label{sec:evaluation}




The evaluation results in terms of \emph{safety} are shown in Table~\ref{table-author-obfuscation-results}. 
The table compares how well each of the three systems that participated in the PAN-2016 Author Obfuscation task can fool various author identification systems. A total of 44 authorship verification systems were used, which were submitted to the previous three shared tasks on Authorship Identification at PAN-2013, PAN-2014, and PAN-2015. 
We can see that the output of our PAN-2016 system caused the performance for these 44 systems to drop the most for each of the three years; this means that it performs best in terms of \emph{safety}.

While our PAN-2016 method is effective in terms of \emph{safety}, it could not always produce text that is grammatically correct and contextually inconspicuous.
That is why we introduced the use of transformation magnitude, which we evaluate below. 
For the evaluation, we use the data provided in the PAN-2016 Author Obfuscation task \cite{potthast:2016}. The data consists of 205 documents that have to be obfuscated.\footnote{\url{http://pan.webis.de/clef16/pan16-web/author-obfuscation.html}}

Table~\ref{table:measures-custom-transformations} shows the impact of the obfuscation, using both our PAN-2016 obfuscator and the new one that pays attention to transformation magnitude, on some text metrics.
The first column shows the names of the text metrics. The second column shows the value of the metrics for the input, i.e.,~\emph{before} the obfuscation. Then follow the values after obuscation, when using our \emph{PAN-2016} and our \emph{new} method, respectively. Finally, the values in the \emph{average} column are calculated on the training dataset and on some texts from Project Gutenberg; these are the target values we want to push the metrics towards. We can see that, overall, the obfuscation methods do push the input metrics towards the target. We further see that the PAN-2016 method often overshoots and can push the metric even further away from the target compared to the input value. For example, for \emph{Stop words to word token count ratio}, the input is 0.52, and it is pushed down to 0.45, which is further away from the target of 0.50 than the input was. In several other cases, the push went in the wrong direction, which can be due to the multi-objective optimization -- when changing text to modify one text metric, we change some words, which could affect a number of other metrics. 


\begin{table}[t]
\footnotesize
\centering
\setlength{\tabcolsep}{7pt}
\begin{tabular}{@{ }@{ }lcccc@{ }@{ }}
\bfseries Participant & \bfseries PAN-2013 & \bfseries PAN-2014 EE & \bfseries PAN-2014 EN & \bfseries PAN-2015 \\
\hline
Mihaylova~{\em et al.} \cite{mihaylova:2016} (our) & -0.10 & -0.13 & -0.16 & -0.11 \\ 
Keswani~{\em et al.}~\cite{keswani:2016} & -0.09 & -0.11 & -0.12 & -0.06 \\ 
Mansoorizadeh~{\em et al.}~\cite{mansoorizadeh:2016} & -0.05 & -0.04 & -0.03 & -0.04 \\ 
\hline
\end{tabular}
\caption{\label{table-author-obfuscation-results} \textbf{Evaluation results in terms of \emph{safety}.} We compare the three obfuscation systems that participated in the PAN-2016 Author Obfuscation task. Shown is the average drop in performance for the 44 authorship verification systems submitted to the PAN-2013, PAN-2014, and PAN-2015 Author Identification tasks when running them on the obfuscated vs. the original versions of the test datasets.
}
\end{table}

\begin{table}[tbh]
\centering
\setlength{\tabcolsep}{5pt}
\begin{tabular}{@{ }lcccccc@{ }}
 & \textbf{Before} & \multicolumn{2}{c}{ \bf After Obfuscation} & \textbf{Average} \\
\textbf{Text Metric} & \textbf{(Input)} & \textbf{PAN-2016} & \textbf{New} & \textbf{(Target)} \\
\hline
Punctuation to word token count ratio & 0.14 & 0.14 & 0.15 & 0.15 \\
Uppercase word tokens to all word tokens count ratio & 0.03 & 0.01 & 0.02 & 0.02 \\
Stop words to word token count ratio & 0.52 & 0.45 & 0.50 & 0.50 \\
Word type to token ratio & 0.44 & 0.47 & 0.45 & 0.44\\
Number of nouns & 0.23 & 0.24 & 0.24 & 0.24 \\
Number of adjectives & 0.08 & 0.09 & 0.08 & 0.06 \\
Number of adverbs & 0.07 & 0.09 & 0.08 & 0.07 \\
Number of verbs & 0.20 & 0.21 & 0.21 & 0.19 \\
\hline
\end{tabular}
\caption{\label{table:measures-custom-transformations}\textbf{Impact of the obfuscation on some text metrics.}
The first column shows the name of a text metric. The second column shows the value of the metrics for the input, i.e.,~\emph{before} the obfuscation. Then follow the values after obuscation, when using our \emph{PAN-2016} and our \emph{new} method, respectively.
Finally, the values in the \emph{average} column are calculated on the training dataset and on some texts from Project Gutenberg; these are the target values we want to push the metrics towards.}
\end{table}

\noindent We can further see in Table~\ref{table:measures-custom-transformations} that the new method, which takes transformation magnitude into account, gets much closer to the target compared to our PAN-2016 obfuscation: it matches exactly the values of four out of the eight metrics (compared to just two for the PAN-2016 version), and gets very close to the target for the other four metrics.

Finally, Table \ref{table:obfuscation-examples} looks into \emph{sensibility}. In particular, it illustrates how the original texts are changed by the different systems (our initial PAN-2016 system, our new system that pays attention to transformation magnitude, and the remaining two systems that participated in PAN-2016). We can see from the examples that our methods perform better than the rivaling PAN-2016 systems, and that
\emph{sensibility} improves when using transformation magnitude.
We can further see that the PAN-2016 system that uses machine translation \cite{mansoorizadeh:2016} can alter the semantics of the input text, which often makes it meaningless. Moreover, the PAN-2016 system that uses word substitution \cite{keswani:2016} is too conservative and often does not change the text at all.


\begin{table}[tbh]
\centering
\begin{tabular}{|p{2.4cm}|p{2.4cm}|p{2.4cm}|p{2.4cm}|p{2.4cm}|} 
\hline
\textbf{Original Text} & \textbf{Machine Translation} \cite{keswani:2016} & \textbf{Word Substitution} \cite{mansoorizadeh:2016} & \textbf{Our PAN-2016 Obfuscation} \cite{mihaylova:2016} & \textbf{Our New Obfuscation}\\
\hline
I am proud. Though I carry my love with me to the tomb, he shall never, never know it.
&
I believe expensive Though continue to never, tomb, it. ever be learned
&
Though I carry my love with me to the tomb , he shall never , never know it .
&
myself 'm proud in them, and though myself carry my beloved with me to the tomb he shall ever ever know it.
&
I 'm proud of them; and though I carry my beloved with me to the tomb he shall ever ever know it.
\\
\hline
4) Religion discriminates. Sure, it unifies (...). On the other hand
&
4) religion discriminates.
some people Sure, unifies (...)
second , it condems
&
4 ) Religion discriminates .
Sure , it unifies (...)
On the other hand
&
Four) Religion discriminate; as sure, it unified (..), and on the other hand
&
Four) Religion discriminate, sure, it unified (...); on the other side
\\
\hline
Consequently, they see a connection between development in spiritual life and professional economic development
&
they Consequently, a link between the spiritual development and economic professional development.
&
Consequently , they see a connection between growth in spiritual life and professional economic development .
&
Definitely, Consequently, they see a connection between development inside spiritual life also professional economic development;
&
Consequently, they see a link between development in spiritual life and professional economic development,
\\
\hline
\end{tabular}
\caption{\label{table:obfuscation-examples}\textbf{Obfuscation examples.} Shown are examples of how the different systems that participated in the PAN-2016 Author Obfusction task transform the original text; the last column shows the output of our new obfuscation method, which we introduced in this paper.}
\end{table}

Overall, we can conclude that our method with transformation magnitude is promising and performs well (better than the three systems that participated in the PAN-2016 Author Obfuscation task) in terms of \emph{sensibility} and \emph{soundness}. In future work, we need to study how it performs in terms of \emph{safety}.

\section{Conclusion and Future Work}
\label{sec:future}

We have described our mediocrity approach to style masking and author obfuscation, which changes the text, so that it is pushed towards average values for some general stylometric characteristics, thus making these characteristics less discriminative.



In future work, we plan experiments with a richer set of features as well as with deep learning. We further aim to design an evaluation measure targeting soundness, which would be enabling for this kind of research.
Finally, we plan to use the techniques used in this paper for author imitation. One key difference will be that the goal for the transformations should not be the average metrics, but the metrics for the target author who should be imitated.



\vspace{-2pt}
\section*{Acknowledgments} 
\vspace{-2pt}

We thank the anonymous reviewers for their constructive comments, which have helped us improve the quality of the present paper.

This research was performed by a team of students from MSc programs in Computer Science in the Sofia University ``St Kliment Ohridski''.
The work is supported by the NSF of Bulgaria under Grant No.: DN 02/11/2016 - ITDGate.


\bibliographystyle{splncs03}
\begin{raggedright}
\bibliography{bib}
\end{raggedright}


\end{document}